\documentclass[acmtog]{acmart}
\usepackage{multirow}
\AtBeginDocument{%
  \providecommand\BibTeX{{%
    \normalfont B\kern-0.5em{\scshape i\kern-0.25em b}\kern-0.8em\TeX}}}

\setcopyright{acmcopyright}
\copyrightyear{2024}
\acmYear{2024}
\setcopyright{acmlicensed}\acmConference[SA Conference Papers '24]{SIGGRAPH Asia 2024 Conference Papers}{December 3--6, 2024}{Tokyo, Japan}
\acmBooktitle{SIGGRAPH Asia 2024 Conference Papers (SA Conference Papers '24), December 3--6, 2024, Tokyo, Japan}
\acmDOI{10.1145/3680528.3687616}
\acmISBN{979-8-4007-1131-2/24/12}

\begin{document}

% Title portion
\title{Pano2Room: Novel View Synthesis from a Single Indoor Panorama}

\author{Guo Pu}
\email{guopu@pku.edu.cn}
\affiliation{%
  \institution{Wangxuan Institute of Computer Technology, Peking University}
  \city{Beijing}
  \country{China}}

\author{Yiming Zhao}
\email{zhaoym@pku.edu.cn}
\affiliation{%
  \institution{Wangxuan Institute of Computer Technology, Peking University}
  \city{Beijing}
  \country{China}}

\author{Zhouhui Lian}
\authornote{Corresponding author}
\email{lianzhouhui@pku.edu.cn}
\affiliation{%
  \institution{Wangxuan Institute of Computer Technology, Peking University}
  \city{Beijing}
  \country{China}}

\renewcommand{\shortauthors}{Pu, et al.}

\begin{abstract}
Recent single-view 3D generative methods have made significant advancements by leveraging knowledge distilled from extensive 3D object datasets. However, challenges persist in the synthesis of 3D scenes from a single view, primarily due to the complexity of real-world environments and the limited availability of high-quality prior resources.
In this paper, we introduce a novel approach called Pano2Room, designed to automatically reconstruct high-quality 3D indoor scenes from a single panoramic image. These panoramic images can be easily generated using a panoramic RGBD inpainter from captures at a single location with any camera.
The key idea is to initially construct a preliminary mesh from the input panorama, and iteratively refine this mesh using a panoramic RGBD inpainter while collecting photo-realistic 3D-consistent pseudo novel views. Finally, the refined mesh is converted into a 3D Gaussian Splatting field and trained with the collected pseudo novel views. This pipeline enables the reconstruction of real-world 3D scenes, even in the presence of large occlusions, and facilitates the synthesis of photo-realistic novel views with detailed geometry.
Extensive qualitative and quantitative experiments have been conducted to validate the superiority of our method in single-panorama indoor novel synthesis compared to the state-of-the-art. Our code and data are available at \url{https://github.com/TrickyGo/Pano2Room}.

\end{abstract}

\begin{CCSXML}
<ccs2012>
   <concept>
       <concept_id>10010147.10010371.10010372</concept_id>
       <concept_desc>Computing methodologies~Rendering</concept_desc>
       <concept_significance>500</concept_significance>
       </concept>
   <concept>
       <concept_id>10010147.10010371.10010382.10010385</concept_id>
       <concept_desc>Computing methodologies~Image-based rendering</concept_desc>
       <concept_significance>500</concept_significance>
       </concept>
 </ccs2012>
\end{CCSXML}

\ccsdesc[500]{Computing methodologies~Rendering}
\ccsdesc[500]{Computing methodologies~Image-based rendering}


\keywords{Image-based Rendering, image-based modeling, texture synthesis and inpainting}

\begin{teaserfigure}
\centering
\resizebox{1.0\linewidth}{!}{
  \includegraphics[width=\linewidth]{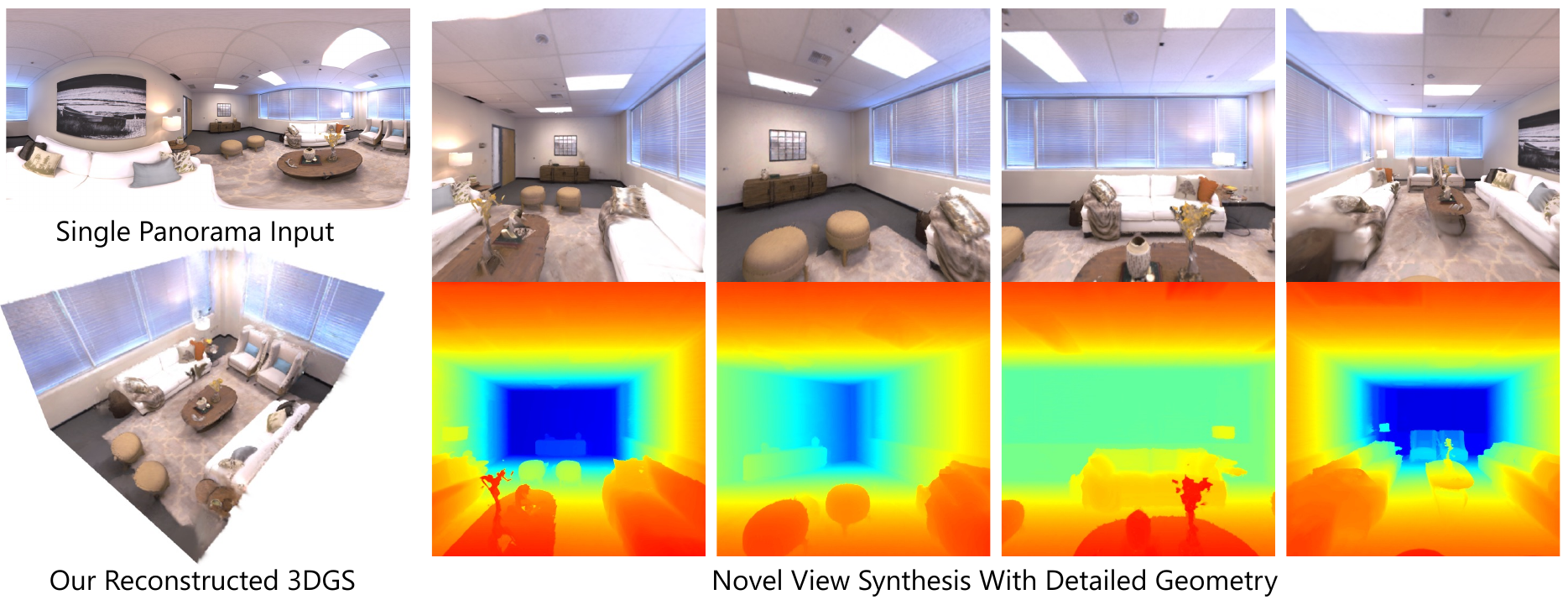}}
  \vspace{-0.4cm}
  \caption{With a single panorama as input, the proposed Pano2Room automatically reconstructs the corresponding indoor scene with a 3D Gaussian Splatting field, capable of synthesizing photo-realistic novel views as well as high-quality depth maps. The panorama is generated using our panoramic RGBD inpainter based on any capture at a single location easily acquired by an average user. For better visualization of the 3D scene in all figures, Gaussian points or mesh blocking the room interior are deleted.}
  \label{fig:Teaser}
  \vspace{0.2cm}
\end{teaserfigure}

\maketitle

\section{Introduction}
\label{sec:intro}

Generating immersive 3D experiences is a crucial task in computer graphics, offering extensive practical applications in fields like Augmented Reality and Virtual Reality. While recent multi-view 3D reconstruction methods such as 3DGS (3D Gaussian Splatting)~\cite{kerbl20233d}, NeuralRecon~\cite{sun2021neuralrecon}, and GFLF~\cite{yin2023geometry} have made significant progress in creating high-quality real-world scenes, they typically require a large number of images to generate accurate 3D representations for complex indoor scenes. The reliance on extensive image collection processes can be both costly and time-consuming, thereby limiting their wider practical applications.
On the other hand, although single-view 3D generative methods such as RealFusion~\cite{melas2023realfusion}, One-2-3-45~\cite{liu2024one}, and DreamGaussian~\cite{tang2023dreamgaussian} have made significant advancements in synthesizing 3D objects by distilling knowledge from extensive 3D object datasets, the task of single-view 3D indoor scene synthesis remains a tough challenge. This is primarily due to the complexity of real-world environments with open-vocabulary objects (objects with undefined/unrecognized semantics) and the limited availability of high-quality prior sources.
In this paper, we present Pano2Room, a novel view synthesis method that leverages multiple priors to achieve the reconstruction of a complete 3DGS for indoor scenes with only a single panorama, as opposed to existing methods that rely on multi-view and multi-location captures.

%Challenges
Reconstructing the complete 3D scene from a single panorama presents significant challenges, as it requires interpreting the complex hidden structures within extremely limited 3D conditions offered by the single panoramic image. These challenges arise from the need to generate a 3D scene that not only faithfully preserves user-captured content and accurately reconstructs the geometry but also infers large-size occluded content with detailed 3D-consistent textures and geometry, integrating them consistently with existing scenes.
Due to these challenges, existing single-view scene reconstruction methods, such as Text2Room~\cite{hollein2023text2room}, PERF~\cite{wang2023perf} and others, struggle not only to create 3D scenes that accurately preserve user-captured content but also to handle large-size occlusion scenarios.

To address these issues, we present Pano2Room, a novel view synthesis method that accurately preserves user-provided captures while generating 3D-consistent new textures and geometry within existing scenes. Specifically, we first convert the input panorama into a mesh and iteratively refine the mesh by leveraging a panoramic RGBD inpainter to generate occluded content and geometry while gradually incorporating the new content into the inpainted mesh. Finally, the inpainted mesh is converted to a 3DGS and trained with collected 3D-consistent pseudo novel views. The optimized 3DGS is capable of synthesizing photo-realistic novel views with detailed geometry.

In summary, the major contributions of this paper are threefold:

1. We introduce Pano2Room, a novel view synthesis method that generates a 3DGS from a single panorama. To the best of our knowledge, Pano2Room is the first work capable of generating a complete 3DGS from a single panorama.

2. We propose a series of new modules specifically designed to improve performance and handle large-size occlusions, including a Pano2Mesh module with improved mesh filtering, a panoramic RGBD inpainter with improved inpainting quality and surface geometry, an iterative mesh refinement module with camera searching and geometry conflict avoidance strategy, and a Mesh2GS module to boost novel view synthesis quality.

3. We extensively evaluate the proposed Pano2Room on various challenging datasets, validating the state-of-the-art novel view synthesis quality in the single-panorama novel view synthesis task.

\section{Related Work}

\subsection{Single Image Novel View Synthesis}

Single-image novel view synthesis approaches such as SynSin~\cite{wiles2020synsin}, PNVS~\cite{xu2021layout}, InfiniteNature~\cite{liu2021infinite} and InfiniteNature-Zero~\cite{li2022infinitenature} address the challenge of single image novel view synthesis without interpreting each scene with explicit 3D representations, thus lacking 3D-consistency in the synthesized novel views. Incorporating explicit 3D representations, layer-based methods like SLIDE~\cite{jampani2021slide}, 3D-Photography~\cite{shih20203d}, AdaMPI~\cite{han2022single} and SinMPI~\cite{pu2023sinmpi} represent a 3D scene using discrete layers, allowing them to generate high-quality synthesis results from a single input image. However, these methods are unsuitable for 360-degree scenes due to the structural limitations of the layered representations.
Recent notable advancements in single-image 3D object generation, such as RealFusion~\cite{melas2023realfusion}, SyncDreamer~\cite{liu2023syncdreamer}, DreamGaussian~\cite{tang2023dreamgaussian}, and One-2-3-45~\cite{liu2024one} distill knowledge from large-scale 3D object datasets, but they are not applicable to indoor scenes due to the complexity of real-world environments with numerous open-vocabulary objects. 

Approaches based on NeRF~\cite{mildenhall2021nerf} and 3DGS~\cite{kerbl20233d} have demonstrated remarkable results in rendering novel views. Several single-view novel view synthesis methods have emerged, including NerfDiff~\cite{gu2023nerfdiff}, NerDi~\cite{deng2023nerdi}, PixelNerf~\cite{cai2022pix2nerf}, DietNeRF~\cite{jain2021putting}, Pix2NeRF~\cite{cai2022pix2nerf}, SinNeRF~\cite{xu2022sinnerf}, OmniNeRF~\cite{hsu2021moving}, PERF~\cite{wang2023perf}, PixelSplat~\cite{charatan2024pixelsplat}, RealmDreamer~\cite{shriram2024realmdreamer} and LucidDreamer~\cite{chung2023luciddreamer}.
Currently, PERF~\cite{wang2023perf} stands as the state-of-the-art single-panorama novel view synthesis method. PERF trains a NeRF with novel views synthesized through a collaborative RGBD inpainting method, enabling high-quality NeRF rendering. However, due to the under-fitting of the NeRF with very few training views and dense sequential inpainting camera trajectory with possible undesired inpainting context and error accumulation, occluded areas in the novel views of PERF are prone to over-smoothed textures and meaningless geometry. 
Based on a single image or text prompts, LucidDreamer~\cite{chung2023luciddreamer} trains a 3DGS by generating pseudo novel views through point cloud rendering. However, unlike meshes, point cloud rendering lacks surfaces, resulting in 3D-inconsistent pseudo novel views and heavy ghost artifacts in the trained 3DGS, hence failing to apply to indoor scenarios with occlusions. 

\subsection{Indoor Novel View Synthesis}

NeuralRoom~\cite{yang2022neural} learns prior knowledge of objects via an offline stage and then synthesizes the room with unseen furniture arrangement. RoomDreamer~\cite{song2023roomdreamer} utilizes diffusion models to edit a given mesh based on prompts. 
ControlRoom3D~\cite{schult2024controlroom3d} and CtrlRoom~\cite{fang2023ctrl} create 3D scene meshes based on semantic layout proxies. While they are capable of creating virtual scenes, they do not
reconstruct real scenes from user captures. These methods cannot be directly applied with only a panorama available.

From a single panorama, Auto3DIndoor~\cite{yang2018automatic} utilizes geometric and semantic cues to recover indoor room layouts and typical indoor objects. DeepPanoContext~\cite{zhang2021deeppanocontext} estimates layouts and object poses, then reconstructs the scene with objects. Pano2CAD~\cite{xu2017pano2cad} estimates the geometry of a room and the 3D poses of objects from a single panorama.  However, in real-world captures, open-vocabulary object detection and generation are challenging and error-prone, leading to these methods to fail in reconstructing complete real-world scenes with occlusions.

Text2Room~\cite{hollein2023text2room} and RGBD2~\cite{lei2023rgbd2} generate 3D scene meshes based on a single input image. Text2Room generates new textures utilizing SD (Stable Diffusion)~\cite{rombach2021highresolution} and predicts new geometries with IronDepth~\cite{bae2022irondepth} in an iterative process to create 3D meshes of rooms. However, the linear depth alignment in Text2Room can result in cracks in surfaces, further leading to error accumulation and poor geometry. Additionally, the edge length filter proposed in Text2Room can unintentionally remove input textures and generate occlusion borders with noticeable irregularities. The reliance on Poisson surface reconstructions~\cite{kazhdan2006poisson} results in over-smoothing artifacts in synthesized novel views.

\section{Method}
\begin{figure*}[t]
\centering
\resizebox{1\linewidth}{!}{
  \includegraphics[width=\linewidth]{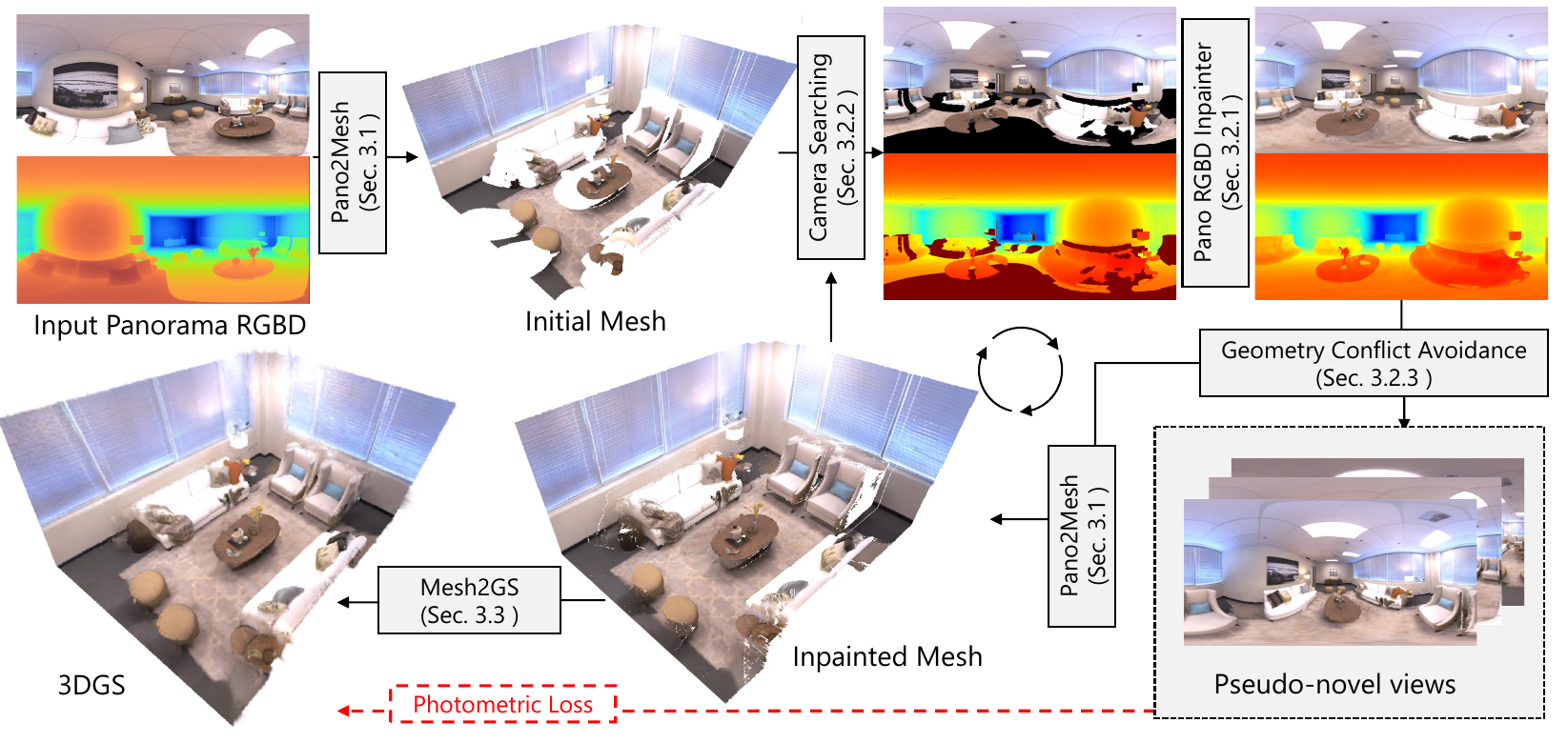}}
  \vspace{-0.3cm}
  \caption{An overview of Pano2Room. With a panorama as input, we first predict the geometry of the panorama using the panoramic RGBD inpainter. Then we synthesize the initial mesh using a Pano2Mesh module. Next, we iteratively search for cameras with the least view completeness, and under the searched viewpoint, we render the existing mesh to obtain panoramic RGBDs with missing areas. To complete each rendered RGBD, we use the panoramic RGBD inpainter to generate new textures and predict new geometries. The new textures/geometries are iteratively fused into the existing mesh if no geometry conflict is introduced. Finally, the inpainted mesh is converted to a 3DGS and trained with collected pseudo novel views.}
  \label{fig:Overview}
  \vspace{-0.0cm}
\end{figure*}

With a single panorama as input, Pano2Room reconstructs the corresponding 3DGS Scene and enables high-quality novel view synthesis. The overview of Pano2Room is shown in Fig.~\ref{fig:Overview}. Given the input panoramic image $I$ and its depth map $D$ (predicted or converted from Cubemaps depth captured by standard RGBD cameras or Lidars), we first generate the initial mesh $Mesh_{init}$ using a Pano2Mesh module (Sec. 3.1). Subsequently, we iteratively refine the mesh into an inpainted mesh $Mesh_{inp}$ (Sec. 3.2). Finally the inpainted mesh is converted to a 3DGS $GS$ (Sec. 3.3).

Concretely, $D$ is captured or predicted by our panoramic RGBD inpainter (Sec. 3.2.1). During the iterative mesh refinement, we search iteratively for optimal camera viewpoints (3.2.2), generate occluded texture and geometry using the panoramic RGBD inpainter (Sec. 3.2.1), and we apply geometry conflict avoidance (Sec. 3.2.3) for generated content while collecting pseudo novel views. Following initialization, $GS$ is optimized with the collected pseudo novel views.

\subsection{Pano to Mesh}

\begin{figure}[t]
\centering
\resizebox{\linewidth}{!}{
  \includegraphics[width=1\linewidth]{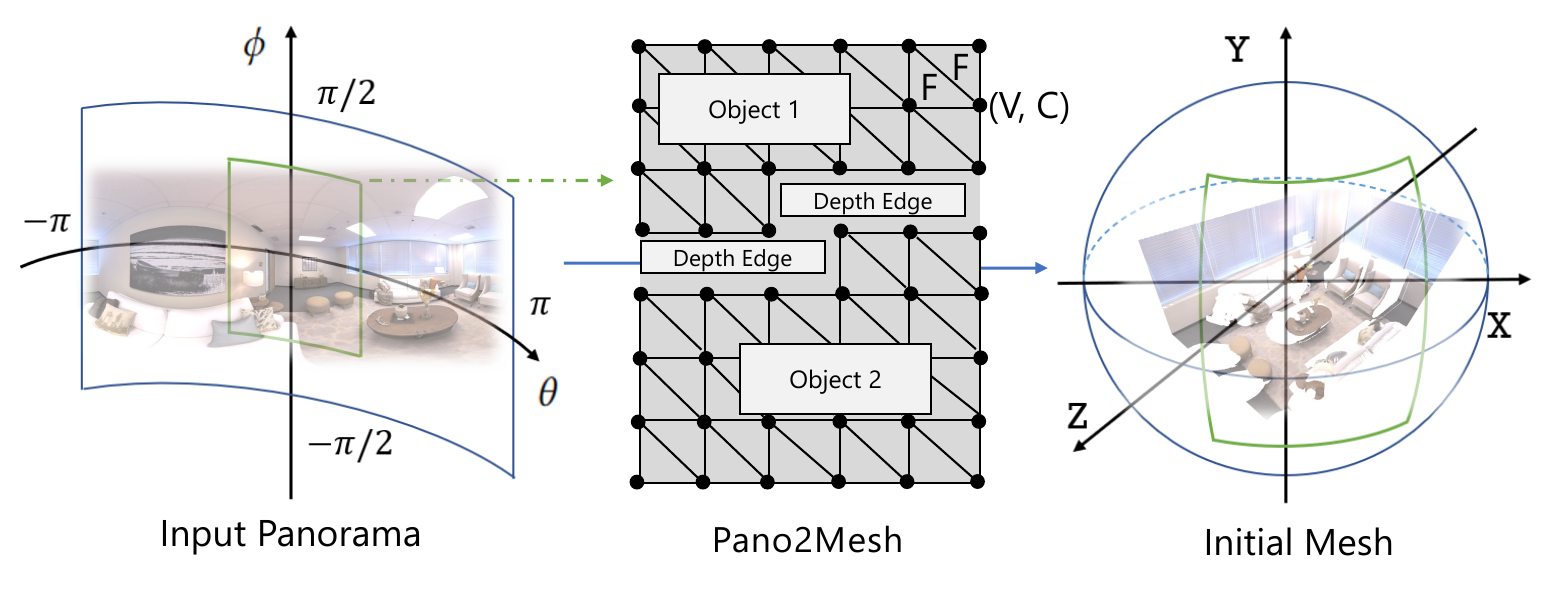}}
  \vspace{-0.3cm}
  \caption{Demonstration of how to convert a panorama to a mesh. Initial mesh vertices and colors are derived from the input panorama’s pixels (depicted as black dots). Triangulation connects neighboring vertices to form faces (depicted by black lines). The edge map of the depth is utilized to disconnect faces representing different objects, ensuring accurate mesh generation.
 }
  \label{fig:PanoToMesh}
  \vspace{-0.2cm}
\end{figure}

Generating 3D-consistent novel views from a single image necessitates dense and coherent surface rendering, challenging for point clouds, NeRF, or 3DGS, which demand equally dense captures and modeling.
Inspired by Text2Room~\cite{hollein2023text2room}, we generate the mesh of the input panorama by first performing triangulation among neighboring pixels in the image space and back-projecting the pixels to the 3D space. In this manner, we can faithfully preserve the content of the input image and enable novel view synthesis with 3D-consistent surfaces. 

Specifically, we convert the input panorama $I$ into an initial mesh $Mesh_{init} = \left \{V, C, F  \right \} $ where $V, C, F$ represent vertices, colors and faces of the mesh, respectively. 
As depicted in Fig.~\ref{fig:PanoToMesh}, we initialize $V$ and $C$ using the screen coordinates $(u,v)$ and corresponding colors from the input panorama, respectively. 
Subsequently, we triangulate $V$ based on $(u,v)$, connecting every four neighboring vertices in a grid to create two faces. This triangulation process yields a mesh connecting all adjacent points.

To disconnect vertices belonging to different objects and eliminate faces with excessive stretching, Text2Room~\cite{hollein2023text2room} utilizes edge length thresholds to remove $F$ with edge lengths surpassing a predefined threshold. However, notice that the scale of the faces is proportional to the depth due to the back-projection from image to 3D space, this strategy tends to unintentionally eliminate content from the input panorama or the inpainted content situated at a large distance and also fails to disconnect objects that are close in proximity. This is because the edge length threshold is scale-dependent, whereas the triangulation process generates object surfaces with varying face scales according to their distances from the camera.

To mitigate this issue, we propose a depth edge filter that employs an edge detector to extract the depth edge mask $M_{D}$ from the depth map $D$. Subsequently, the vertices within $M_{D}$ are interpreted as the silhouettes of objects and are disjointed. This filter effectively filters out unwanted faces in a scale-invariant manner without compromising the input textures. Consequently, it enhances the smoothness of edges in occluded regions, leading to significant improvements in the results of the image inpainter. Illustrations demonstrating the effectiveness of the depth edge filter are provided in the supplemental materials.

Following mesh filtering in the image space, we convert the vertices from the image space $(u, v)$ to the 3D space $(x, y, z)$ based on the depth map $D$ using spherical projection:
\begin{equation}
\begin{aligned}
\phi &= v/H \cdot \pi,  \theta = u/W \cdot 2\pi - 0.5\pi, \\
x &= \sin(\phi ) \cdot \cos(\theta) \cdot D(\phi,\theta),\\
y &= \cos(\phi) \cdot D(\phi,\theta),\\
z &= \sin(\phi ) \cdot \sin(\theta) \cdot D(\phi,\theta),\\
\end{aligned} 
\end{equation}
where $W$ and $H$ represent the width and height of $I$ respectively, and  $(\theta, \phi)$ denote the polar coordinates of $(u, v)$, respectively. Through the above projection, we obtain an initial mesh $Mesh_{init}$.

\subsection{Iterative Mesh Completion}

We propose an iterative refinement method for the initial mesh $Mesh_{init}$ by integrating the generated occluded content into the existing mesh. The occluded textures are produced through image inpainting using pre-trained image generators. In the image inpainting task, having more context often aids the inpainter in generating new content that is consistent with the existing content. Since panoramas contain complete context in a single location, we sample panoramic views for inpainting instead of perspective views, which have a limited field of view and less context.

In each iteration, we start by identifying a camera viewpoint $Cams_{i}$ with the least view completeness (Sec. 3.2.2). We then render the existing mesh using a mesh renderer $R_{Mesh}$ with a vertex color shader to obtain the rendered panorama image $I_{render}$, mask $M_{render}$, and depth map $D_{render}$. Next, we employ a panoramic RGBD inpainter to generate new textures $I_{inp}$ and predict new geometry $D_{inp}$ for $I_{inp}$. 
Finally, we generate a mesh for the newly generated content $I_{inp} * (1-Mask_{render})$ using the Pano2Mesh module (Sec. 3.2) and merge this new mesh into the existing mesh.
After $N_{mesh}$ iterations, we obtain the inpainted mesh $Mesh_{inp}$. Further details are provided in the subsequent subsections.

\subsubsection{Panoramic RGBD Inpainter}

To synthesize the occluded content, we propose a panoramic RGBD inpainter including a panoramic image inpainter that generates high-quality textures consistent within the existing scene and a panoramic depth inpainter that predicts detailed new geometries aligned with existing surfaces.

\begin{figure*}[t]
\centering
\resizebox{1\linewidth}{!}{
  \includegraphics[width=\linewidth]{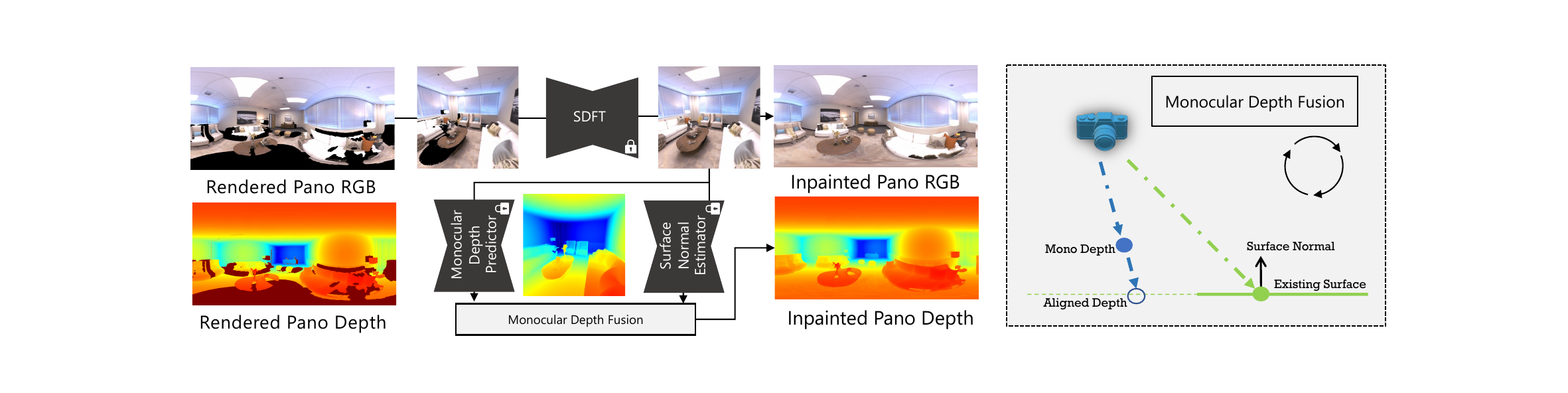}}
  \vspace{-0.3cm}
  \caption{Panoramic RGBD Inpainter. We first inpainted the rendered panoramic image using SDFT. Then, the depth of inpainted content is estimated by a pre-trained monocular depth predictor and seamlessly fused into the rendered panoramic depth, creating inpainted panoramic depth with detailed new geometry aligned with existing geometries and enforced surface normals.}
  \label{fig:PanoRGBDInpainter}
  \vspace{-0.0cm}
\end{figure*}

As depicted in Fig.~\ref{fig:PanoRGBDInpainter}, with the rendered panoramic image $I_{render}$ and its depth $D_{render}$ as input, the panoramic RGB inpainter generates the inpainted panoramic image $I_{inp}$ and the panoramic depth inpainter predicts the inpainted depth $D_{inp}$. 

\paragraph{Panoramic Image Inpainter}

Existing panorama inpainting methods face limitations in generating high-quality content, particularly at high resolutions. For example, diffusion-based techniques such as MultiDiffusion~\cite{bar2023multidiffusion} tend to produce pseudo-panoramic images that do not strictly conform to the spherical projection relationship.

To address these limitations, we inpaint $I_{render}$ by leveraging the powerful image generation model SD (Stable Diffusion)~\cite{rombach2021highresolution}, which uses semantic masks as the conditional input. The conditional latent diffusion model is trained by minimizing:
\begin{equation} \begin{aligned}
    L_{LDM} := \mathbb{E}_{\varepsilon(x),y,\epsilon\sim \mathcal{N}(0,1),t}[\left\| \epsilon-\epsilon_{\theta}(z_{t},t,m) \right\|^{2}_{2}], \label{eq:diff}
\end{aligned} \end{equation}
where $t = 1, 2, ... T$ denotes the time step, $z_{t}$ is the noisy version of the latent vector $z$ of the input $x$, $m$ represents the mask, $\epsilon$ is the noise schedule, and $\epsilon_{\theta}$ denotes the time-conditioned U-Net, respectively.

As SD takes perspective images as input rather than panoramic images, we iteratively sample $k \in K$ perspective images $I_{render}(k)$ from $I_{render}$, employ SD to generate an inpainted perspective image $I_{inp}(k)$, then warp $I_{inp}(k)$ back to $I_{render}$. This iterative process ensures that each sampled $I_{render}(k)$ is conditioned on the existing panoramic image, which includes all previously inpainted content.
Formally, the tangent projection images $I_{render}(k)$ is produced by the $K=20$ faces of an icosahedron that uniformly covers the sphere’s surface. 

\begin{figure}[t]
\centering
\resizebox{\linewidth}{!}{
  \includegraphics[width=\linewidth]{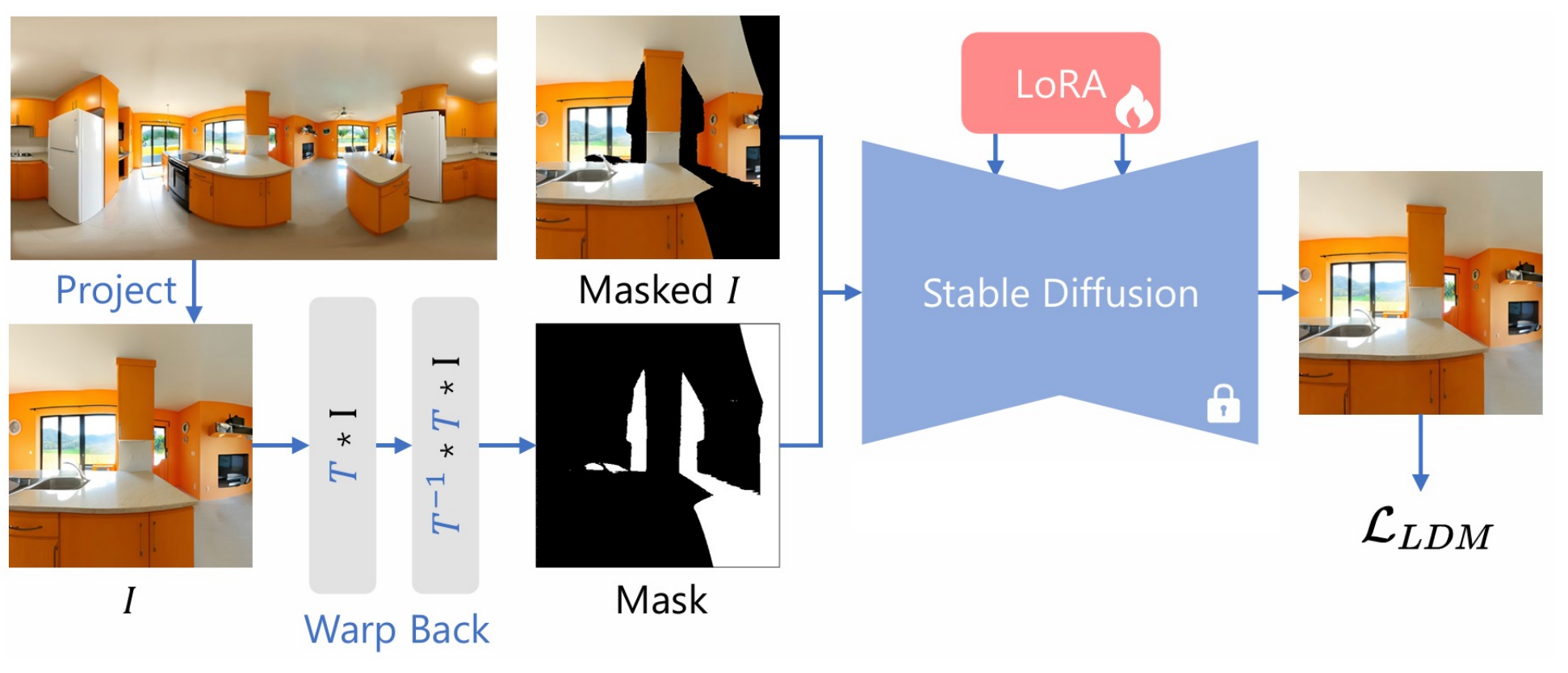}}
  \vspace{-0.3cm}
  \caption{SDFT: Fine-tuning Stable Diffusion on the input panorama to learn the styles and features.}
  \label{fig:IndoorImageInpainter}
  \vspace{-0.3cm}
\end{figure}

However, using perspective images as input, SD cannot capture the full style of the scene, leading to inconsistent generated new textures. To faithfully capture the style and features of the input panorama, we propose to fine-tune SD (SDFT) for each panorama. 
Inspired by SinMPI~\cite{pu2023sinmpi}, we create pseudo inpainting pairs by employing monocular mesh construction and projection to fine-tune SD for each panorama, as depicted in Fig.~\ref{fig:IndoorImageInpainter}.

% finetune lora
In pursuit of rapid convergence and parameter efficiency, we employ Low-rank Adaptation (LoRA)~\cite{hu2021lora} to fine-tune the inpainting model for each individual scene. Specifically, we maintain SD in a frozen state and only update the low-rank matrices of the self-attention layers within the U-Net during the training process. The optimization objective remains consistent with Eq.~\ref{eq:diff}. We optimize the parameters of the self-attention layers due to their critical role in image inpainting tasks. Relevant ablation study results provided in supplemental materials show that the SDFT inpainter is able to capture the style and features of the input panorama, and generate new textures that are more consistent with the scene. 

Taking advantage of the powerful image generation capabilities of SDFT, completing the inpainting process ensures a continuous and seamless inpainted panoramic image preserving the original resolution with high-quality inpainting.

\paragraph{Panoramic Depth Inpainter}

The depth of a panorama includes visible surfaces in an indoor scene, incorporating complex textures and geometries with intricate details. Following 360MonoDepth~\cite{rey2022360monodepth}, to inpaint $D_{inp}$ based on $D_{render}$, we first predict the depth maps $D_{inp}(k)$ from multiple perspective projects $I_{inp}(k)$ of $I_{inp}$ using a pretrained monocular depth estimator DPT~\cite{ranftl2021vision}, and then fuse these $D_{inp}(k)$ to $D_{render}$ to produce $D_{inp}$.

To iteratively fuse $D_{inp}(k)$ into $D_{render}$, we optimize the re-scale factors of $D_{inp}(k)$ to obtain a complete, smooth inpainted depth $D_{inp}$ with detailed geometry. Specifically, $D_{inp}$ is initialized as the pixel-wise average of the spherical projection of all $D_{inp}(k)$. For each $D_{inp}(k)$, we aim to re-scale $D_{inp}(k)$ by a scaling factor $s(k)$ and an offset $o(k)$ to obtain the optimized depth $\tilde{D}_{inp}(k)=s(k)D_{inp}(k)+o(k)$. The optimization objective is defined as:
 \begin{equation} \begin{aligned}
\mathop{\arg\min}\limits_{{\left \{s(k), o(k)  \right \}}} E_{fix} + E_{align} + E_{normal}+ E_{smooth},
\end{aligned} \end{equation}
where the existing depth is fixed during depth fusion through:
\begin{equation} \begin{aligned}
E_{fix}=\sum_{k\in K}((D_{render}(k)-\tilde{D}_{inp}(k))*M_{render}(k))^{2}.
\end{aligned} \end{equation}
Depth is constrained to be consistent across all tangent views through the alignment loss:
\begin{equation} \begin{aligned}
E_{align}=\sum_{(a,b)\in K}\sum_{x\in\Omega (a,b)}(\tilde{D}_{inp}(a)(x)-\tilde{D}_{inp}(b)(x))^{2},
\end{aligned} \end{equation}
where $a,b$ denote any two $\tilde{D}_{inp}$ with overlapping pixels $x$ from the overlapping regions $\Omega(a,b)$. The spatial smoothness between neighboring grid-points $m$ and $n$ is ensured by applying the smoothness loss:

\begin{equation} \begin{aligned}
E_{smooth}=\sum_{k\in K}\sum_{(m,n)}\left \|s_{k}^{m}- s_{k}^{n} \right \|_{2}^{2} + \left \|o_{k}^{m}- o_{k}^{n}  \right \|_{2}^{2}.
\end{aligned} \end{equation}

In indoor scenes, dense and coherent surface rendering is one of the most critical factors for high-quality novel view synthesis. We propose to enforce surface normal constraints by using pre-trained surface normal estimators and regulating $D_{inp}$ with a surface normal loss:
\begin{equation} \begin{aligned}
E_{normal}=\sum_{k\in K}(N_{I_{inp}}(k)-N_{\tilde{D}_{inp}}(k))^{2},
\end{aligned} \end{equation}
where the surface normals $N_{I_{inp}}$ and $N_{\tilde{D}_{inp}}$ are estimated by pretrained surface normal estimators~\cite{bae2022irondepth}.

After $N_{depth}$ iterations of optimization, the monocular depth maps are seamlessly fused into the rendered panoramic depth, creating an inpainted panoramic depth $D_{inp}$ with detailed new geometry aligned with existing geometry and enforced surface normals.

\subsubsection{Searching Inpainting Viewpoints}

Existing methods such as PERF~\cite{wang2023perf} complete existing scenes through iterative inpainting of occluded content along pre-defined camera trajectories in a sequence. However, inpainting under any inappropriate camera viewpoint with small grazing angles between the view direction and inpainted surface normal, or conducting multiple inpainting steps for any occluded area leads to lower-quality new texture and geometry generation. This deterioration occurs because image inpainters rely on context for filling, but the accumulated errors of image inpainters and depth estimators lead to a decline in the quality of the input context, resulting in inferior filling quality, including cluttered and uneven surfaces. 

Based on this observation, our objective is to minimize the number of cameras and inpainting steps required to complete a scene. 
In each scene, we search for viewpoints candidates as follows: First, we identify the room boundaries according to the panoramic depth map. Following the Atlanta world assumption~\cite{schindler2004atlanta}, the middle line of the depth map signifies the horizontal boundary of the scene, while the distance between the ceiling and floor indicates the vertical boundaries. Subsequently, we uniformly sample in the room space to define all possible cameras.
Next, we propose a camera search strategy aiming to reduce inpainting steps by identifying cameras that cover the largest areas needing completion in each iteration. Specifically, from a pre-defined set of potential camera viewpoints $Cams_{k}$ across the scene that covers the majority areas of the scene, we search for viewpoints with the least view completeness:
\begin{equation} \begin{aligned}
\mathop{\arg\min}\limits_{{\left \{k\in K  \right \}}} \sum_{(u,v)\in (H,W)}^{}M_{render}(k)(u, v),
\end{aligned} \end{equation}
where $(u,v)$ denote pixel coordinates and $H, W$ represent the height and width of $M_{render}$. The $Cams_{k}$ search space are elaborated in supplemental materials.

As illustrated in Fig.~\ref{fig:CameraSearch}, the sequential camera trajectory strategy results in multiple inpainting steps in occluded spaces, yielding blurry inpainting results. In contrast, utilizing searched viewpoints, our method markedly enhances the generation performance of plausible new textures and geometry.

\begin{figure}[t]
\centering
\resizebox{1\linewidth}{!}{
  \includegraphics[width=1\linewidth]{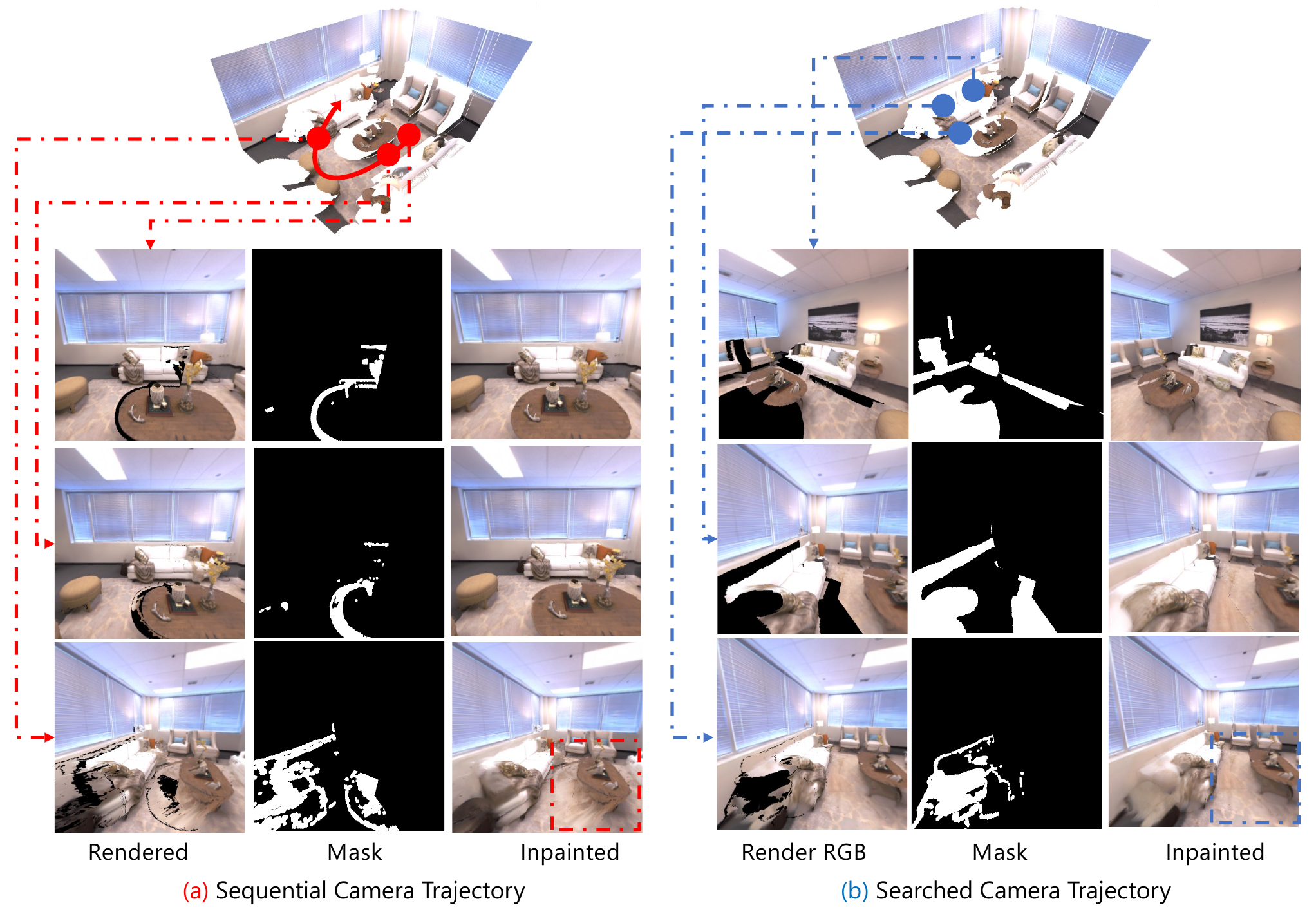}}
  \vspace{-0.2cm}
  \caption{The effectiveness of the proposed camera search strategy. The strategy with sequential camera trajectory leads to multiple inpainting steps performed in an occluded space and produces blurry inpainting results, while our method with searched viewpoints facilitates the generation of plausible new textures and geometry.}
  \label{fig:CameraSearch}
  \vspace{-0.0cm}
\end{figure}

\subsubsection{Geometry Conflict Avoidance}

In each iteration of generating occluded content, newly added content may result in geometric conflicts with previously rendered viewpoints $I_{render}$ or $I_{inp}$, thereby causing 3D-inconsistencies in pseudo novel views. Training 3DGS with 3D-inconsistent images can result in pronounced ghost artifacts and over-smoothing.

In order to avoid geometry conflicts, we propose a fast geometry conflict avoidance strategy based on mesh rendering.
Upon fusing new mesh with the existing mesh, we conduct a render of the mesh from preceding viewpoints to detect any potential geometry conflicts. If any geometry conflicts are introduced, these new faces are abandoned. Specifically, to ensure that newly incorporated mesh does not disrupt previous views, we identify conflicting inpainted content using a mask:
\begin{equation} \begin{aligned}
M_{conflict}(k)(u, v) =  \left\| I_{inp}(k)(u,v) - R_{mesh}(k)(u,v) \right\| > \varepsilon,
\end{aligned} \end{equation}
where $(u,v)$ are the pixel coordinates and $\varepsilon$ stands for the conflict threshold, respectively.

After the inpainting process, we run the Poisson surface reconstruction~\cite{kazhdan2006poisson} to close any remaining holes and obtain a watertight mesh. 

\subsection{Mesh to 3DGS}
The Poisson surface reconstruction~\cite{kazhdan2006poisson} yields a complete mesh, but it tends to create over-smoothing texture and geometry, downgrading photo-realism and sharpness of rendered novel views. To address this problem, we propose to convert the mesh into a 3DGS $GS$ and train $GS$ with collected pseudo novel views $I_{inp}$ to ensure that the rendering quality of $GS$ matches the photo-realism of the images generated by our fine-tuned SD. 

Concretely, we first construct the initial $GS$ using the $V$ and $C$ of the mesh as the initial point cloud. Then, we train $GS$ using perspective views $I_{inp}(k)$ of collected pseudo novel views $I_{inp}$. 
The 3DGS is rendered using point-based alpha-blending, which computes the color $C$ of a pixel by blending $N$ ordered points overlapping the pixel according to:
\begin{equation} \begin{aligned}
C = \sum_{i\in N}^{} c_{i}\alpha_{i}\prod_{j=1}^{i-1}(1-\alpha_{j}),
\end{aligned} \end{equation}
where $c_{i}$ represents the color of each point, and $\alpha_{i}$ is given by evaluating a 2D Gaussian with covariance multiplied with a learned per-point opacity. To be specific, $GS$ is optimized with collected pseudo novel view $I_{inp}(k)$ using photometric losses, comprising an L1 loss combined with a D-SSIM term:
\begin{equation} \begin{aligned}
L = L_{1}(R_{GS}, I_{inp}) +
\lambda L_{D-SSIM}(R_{GS}, I_{inp}),
\end{aligned} \end{equation}
where $R_{GS}$ denotes the 3DGS renderer.

\begin{figure}[t]
\centering
\resizebox{1\linewidth}{!}{
  \includegraphics[width=1\linewidth]{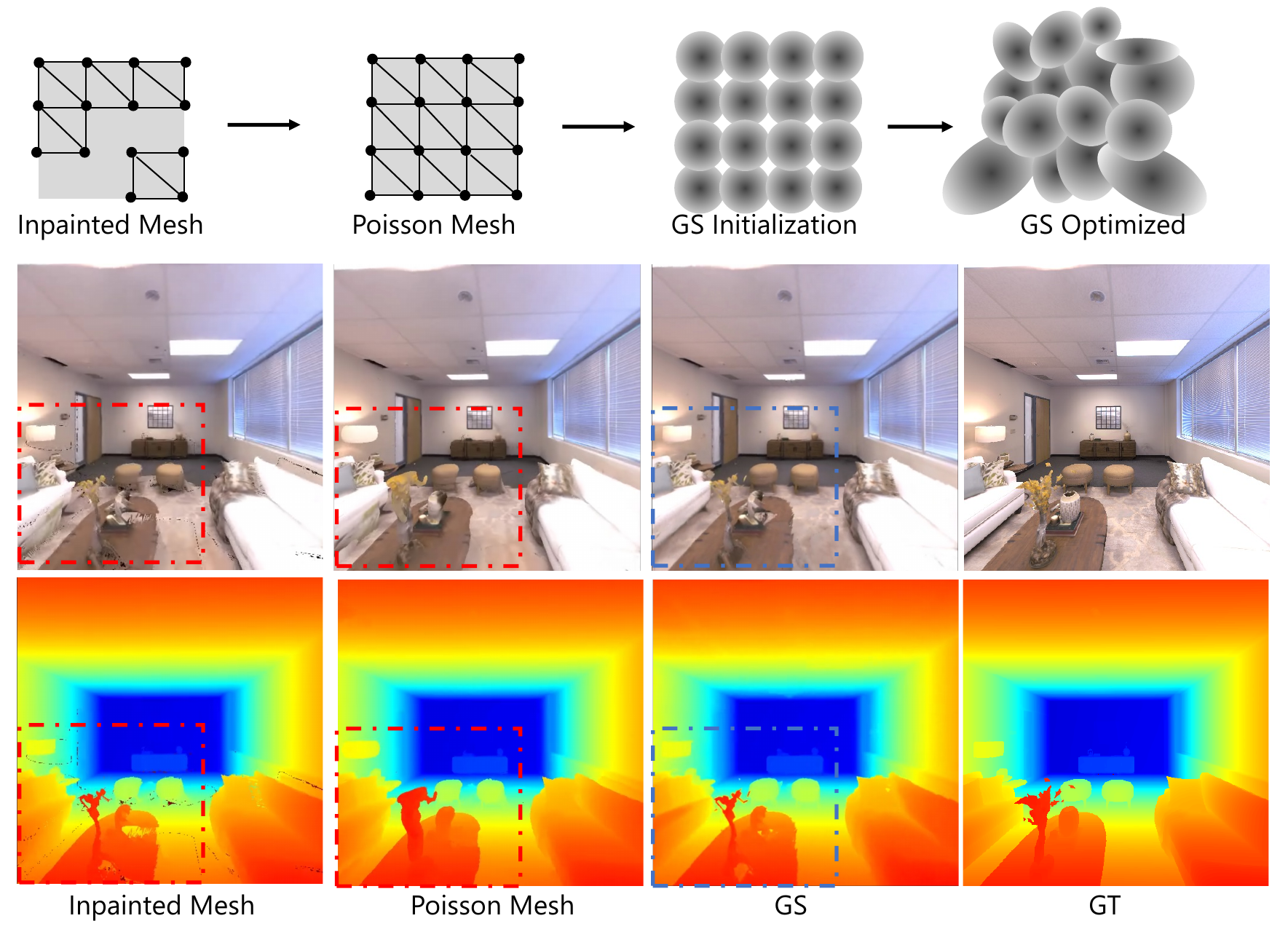}}
  \vspace{-0.2cm}
  \caption{Demonstration of how to convert a mesh to a 3DGS. The optimized 3DGS renders novel views with high-quality and detailed geometry, effectively addressing the over-smoothing artifacts introduced by the Poisson surface reconstruction. Please zoom in for a better inspection.}
  \label{fig:Mesh2GS}
  \vspace{-0.0cm}
\end{figure}

Benefiting from our proposed iteratively mesh completion with the geometric conflict avoidance strategy, the collected pseudo novel views used for training $GS$ are 3D consistent and photo-realistic. This results in an excellent quality of the trained $GS$, significantly enhancing the quality of novel view synthesis.
As illustrated in Fig.~\ref{fig:Mesh2GS}, the optimized $GS$ renders novel views with high quality and detailed geometry, effectively handling the over-smoothing artifacts introduced by the Poisson surface reconstruction. 

\section{Experiments}
\subsection{Comparison with Previous Methods}

\begin{table*}[ht]
\centering
\caption{Quantitative comparison on the Replica dataset.}
\vspace{-0.0cm}
\resizebox{\linewidth}{!}{
\begin{tabular}{l|lll|lll|lll|lll}
\multicolumn{1}{c|}{\multirow{2}{*}{Scene|Method}} & \multicolumn{3}{c|}{Pano2Room}                                                   & \multicolumn{3}{c|}{PERF}                                                        & \multicolumn{3}{c|}{Text2Room}                                                   & \multicolumn{3}{c}{LucidDreamer}                                                \\
\multicolumn{1}{c|}{}                              & \multicolumn{1}{c}{PSNR$\uparrow$} & \multicolumn{1}{c}{SSIM$\uparrow$} & \multicolumn{1}{c|}{LPIPS$\downarrow$} & \multicolumn{1}{c}{PSNR$\uparrow$} & \multicolumn{1}{c}{SSIM$\uparrow$} & \multicolumn{1}{c|}{LPIPS$\downarrow$} & \multicolumn{1}{c}{PSNR$\uparrow$} & \multicolumn{1}{c}{SSIM$\uparrow$} & \multicolumn{1}{c|}{LPIPS$\downarrow$} & \multicolumn{1}{c}{PSNR$\uparrow$} & \multicolumn{1}{c}{SSIM$\uparrow$} & \multicolumn{1}{c}{LPIPS$\downarrow$} \\ \hline
Room 0                                             & \textbf{23.29}           & \textbf{0.788}           & \textbf{0.168}             & 22.58                    & 0.767                    & 0.167                      & 21.55                    & 0.779                    & 0.151                      & 17.50                    & 0.611                    & 0.470                     \\
Room 1                                             & 25.44                    & \textbf{0.886}           & 0.090                      & \textbf{26.10}           & 0.854                    & 0.099                      & 24.92                    & 0.884                    & \textbf{0.079}             & 18.03                    & 0.681                    & 0.484                     \\
Room 2                                             & \textbf{25.06}           & \textbf{0.874}           & 0.127                      & 23.75                    & 0.858                    & \textbf{0.121}             & 22.21                    & 0.862                    & 0.138                      & 17.40                    & 0.719                    & 0.456                     \\
Office 0                                           & \textbf{25.15}           & \textbf{0.891}           & \textbf{0.105}             & 21.15                    & 0.871                    & 0.124                      & 21.68                    & 0.863                    & 0.126                      & 19.25                    & 0.746                    & 0.418                     \\
Office 1                                           & \textbf{31.77}           & \textbf{0.953}           & \textbf{0.043}             & 30.98                    & 0.938                    & 0.070                      & 23.92                    & 0.935                    & 0.078                      & 20.91                    & 0.788                    & 0.332                     \\
Office 2                                           & \textbf{23.22}           & \textbf{0.899}           & \textbf{0.087}             & 20.83                    & 0.874                    & 0.098                      & 20.37                    & 0.885                    & 0.090                      & 14.39                    & 0.678                    & 0.450                     \\
Office 3                                           & \textbf{20.93}           & \textbf{0.858}           & \textbf{0.143}             & 19.69                    & 0.825                    & 0.159                      & 15.77                    & 0.789                    & 0.232                      & 13.98                    & 0.679                    & 0.446                     \\
Office 4                                           & \textbf{26.65}           & \textbf{0.935}           & \textbf{0.064}             & 22.64                    & 0.904                    & 0.086                      & 23.53                    & 0.922                    & 0.065                      & 15.79                    & 0.743                    & 0.415                    
\end{tabular}
}
\label{tab: replica}
\vspace{-0.0cm}
\end{table*}

\subsubsection{Baselines}
We compare our method with three state-of-the-art techniques: PERF~\cite{wang2023perf}, Text2Room~\cite{hollein2023text2room}, and LucidDreamer~\cite{chung2023luciddreamer}. PERF is a single-panorama novel-view synthesis method, Text2Room is a text-to-indoor-mesh generation method, and LucidDreamer is a 3DGS-based single-view novel view synthesis method.
For fair comparison in each scene, all methods are provided with the identical input panoramic RGBD and camera trajectory to align generated novel views with corresponding ground-truth views. Since Text2Room and LucidDreamer take perspective images as input, we project the panoramic RGBD into cubemaps RGBD as their input.

\subsubsection{Datasets}
We evaluate all methods using the eight single-room scenes from the Replica dataset. We establish a consistent rendering camera trajectory comprising 150 poses for each method, with the cameras traversing the room and facing inward. 
Additionally, we also conduct comparisons on real-world captured panoramas in various indoor scenes from the Pano3D dataset~\cite{albanis2021pano3d}, the S2D3D dataset~\cite{2017arXiv170201105A}, and the ZIND dataset~\cite{ZInD}, as detailed in the supplemental materials.

\subsubsection{Evaluation Metrics}
We assess the reconstruction quality of the rendered views produced by each method, evaluating both the fidelity of how faithfully the method preserves the information of the input panorama and the image quality of synthesized novel views.
The rendered views of each method are compared with corresponding ground-truth images using metrics including Peak Signal-to-Noise Ratio (PSNR), Structural Similarity Index (SSIM), and Learned Perceptual Image Patch Similarity (LPIPS).

\subsubsection{Qualitative Comparison}
\begin{figure*}[t]
\centering
\resizebox{1\linewidth}{!}{
  \includegraphics[width=1\linewidth]{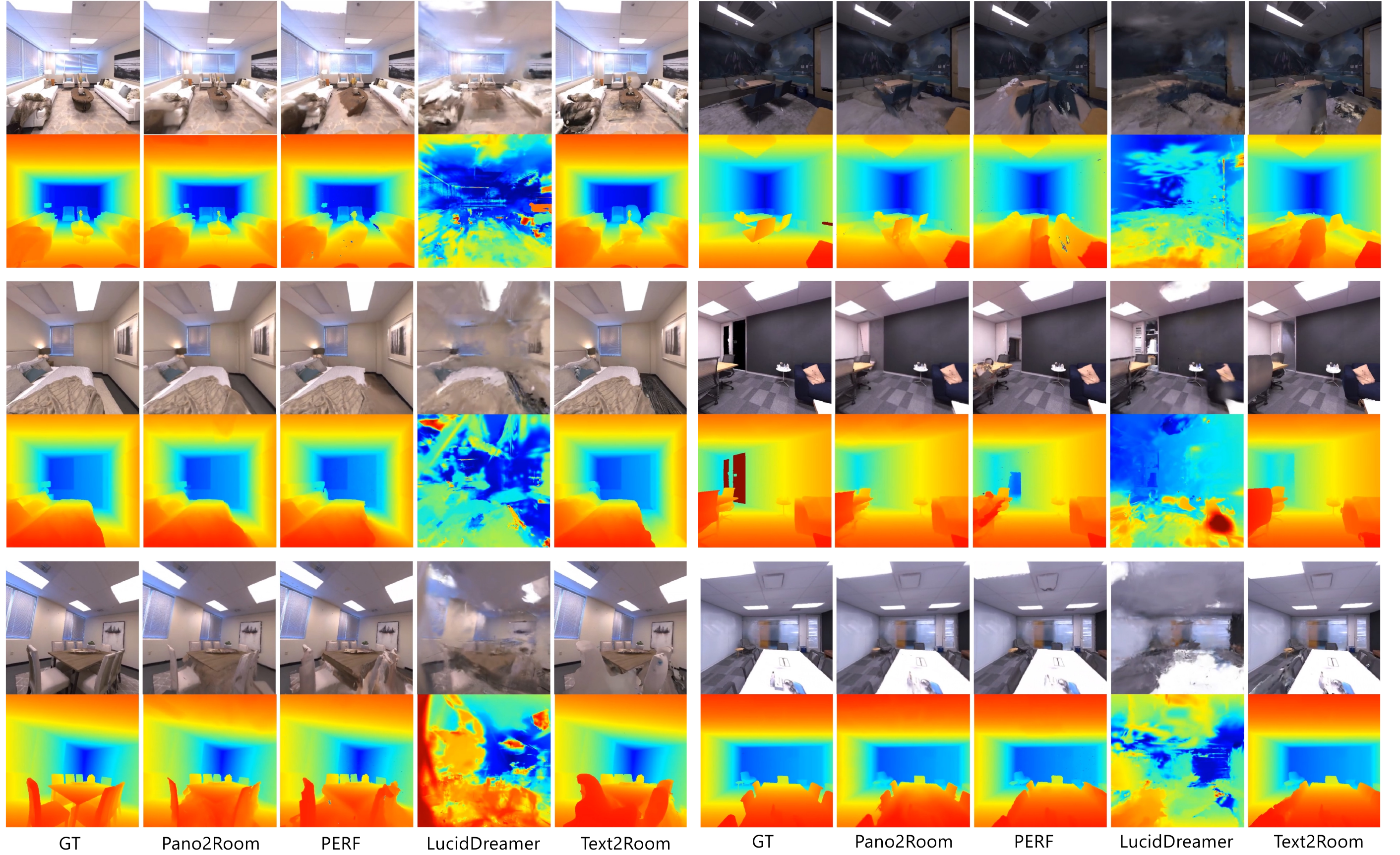}}
  \vspace{-0.0cm}
  \caption{Comparison of novel view synthesis and depth map of each method with the corresponding ground truth. Novel views synthesized by Text2Room and LucidDreamer are prone to strong artifacts. PERF tends to generate over-smoothed occluded areas with interpolated geometries. Our novel views are more photorealistic, considering texture, geometry, and structure. Please zoom in for a better inspection.}
  \label{fig:comparison}
  \vspace{-0.0cm}
\end{figure*}

Fig.~\ref{fig:comparison} shows the qualitative comparison of rendered novel views.
PERF gradually collects pseudo novel-views around the original location, making it challenging to inpaint large areas with appropriate camera positions and contextual information. 
The synthesized novel views of PERF are prone to over-smoothed textures and meaningless interpolated geometry, especially in occluded areas. This is caused by the under-fitting of NeRF with very few training views and dense sequential inpainting camera trajectory with possible undesired inpainting context and error accumulation.
LucidDreamer is prone to strong artifacts due to the fact that point cloud rendering lacks surfaces which means when the camera is close to surfaces, the gaps in the point cloud become apparent. This results in 3D-inconsistent collected novel views for the training of 3DGS, leading to heavy ghost artifacts in the trained 3DGS.
Text2Room suffers from the artifacts brought by Poisson Mesh Reconstruction and undesired inpainting results.
Our novel views demonstrate superior photorealism, exhibiting high-quality texture, geometry, and structural fidelity.
Please refer to the supplemental materials and accompanying video for more comprehensive analysis and visualizations.

\subsubsection{Quantitative Comparison}
Table~\ref{tab: replica} presents the quantitative evaluation metrics on the Replica dataset. Consistent with the visual comparisons, Pano2Room achieves the best performance across most scenes in terms of both reconstruction and novel view quality metrics, validating its state-of-the-art capability in single-panorama novel view synthesis for indoor scenes.
Our method takes approximately 40 minutes to generate a complete 3DGS scene based on a single panorama that can be rendered at 156 FPS, outperforming the state-of-the-art single-panorama novel view synthesis method, PERF. Results of efficiency comparison can be found in supplemental materials.

\subsection{Ablation Study}

%\subsubsection{Full Ablations}

\begin{figure*}[t]
\centering
\resizebox{1\linewidth}{!}{
  \includegraphics[width=1\linewidth]{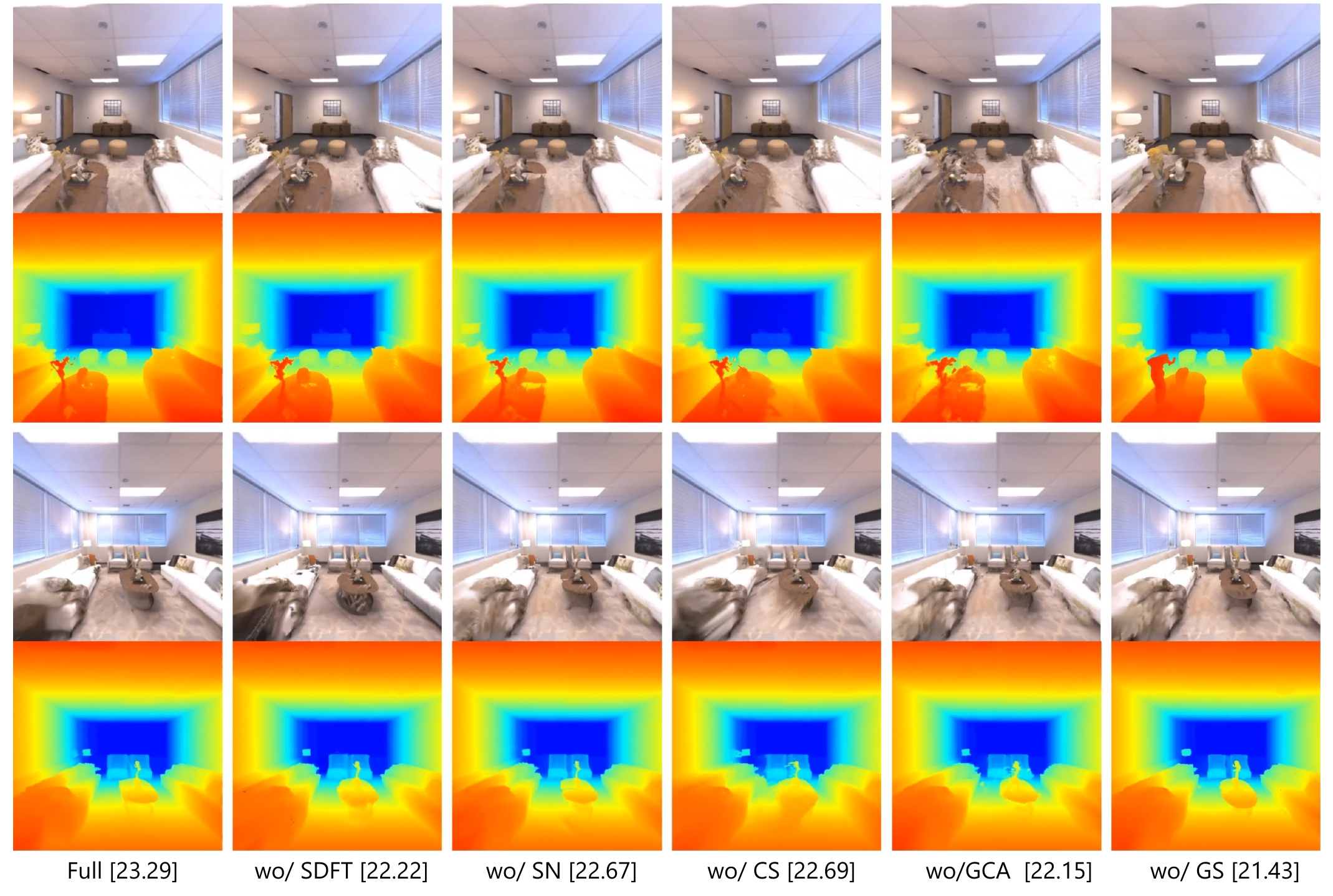}}
  \vspace{-0.0cm}
  \caption{Ablation study results conducted on the representative Room-0 scene from the Replica dataset featuring 150 rendered views and the average PSNR score across all views. We evaluate our method without each key component including Stable Diffusion finetuning (SDFT), Surface Normal Constraints (SN),  Camera Searching strategy (CS), Geometry Conflict Avoidance strategy (GCA), and Mesh2GS module (GS). The corresponding PSNR Scores are provided within the square blankets for quantitative comparison. Please zoom in for a better inspection.}
  \label{fig:AblationFull}
  \vspace{-0.0cm}
\end{figure*}

Fig.~\ref{fig:AblationFull} illustrates the ablation study results of our method conducted on the representative Room-0 scene from the Replica dataset featuring 150 rendered views and the average PSNR score across all views. We evaluate our method without each key component namely Stable Diffusion finetuning (SDFT), Surface Normal Constraint (SN), Camera Searching strategy (CS), Geometry Conflict Avoidance strategy (GCA), and Mesh2GS module (GS).  More specifically, 
(1) \textbf{w/o SDFT}: Employing Stable Diffusion without finetuning, resulting in increased variance in the generated content. This variance leads to a decrease in reconstruction scores due to lower consistency in the generated styles.
(2) \textbf{w/o SN}: Without surface normal constraints, the quality of surface geometry completion decreases, resulting in cracks in surfaces.
(3) \textbf{w/o CS}: Using sequential poses in iterative inpainting causes large occlusion areas to be inpainted from inappropriate camera poses multiple times, leading to low-quality inpainted texture due to error accumulation.
(4) \textbf{w/o GCA}: Without geometry conflict avoidance, geometry conflicts occur where the inpainted new content blocks the user-captured content and previous inpainted content, causing floater artifacts.
(5) \textbf{w/o GS}: The rendering results of the initial 3D mesh are over-smoothed, resulting in the loss of many details from user captures and low-quality novel view synthesis.

\section{Implementation Details}
We performed all experiments utilizing an Nvidia A40 GPU. 
The mesh completion iteration $N_{mesh}$ is adaptive and depends on the occlusion size of the particular scene, typically falling within the range $2 < N_{mesh} < 16$. The monocular depth fusion iterations are set to $N_{depth} = 3000$. The geometry conflict threshold is set to $\varepsilon = 5$. The process of generating the final 3DGS field from a panorama typically takes approximately 40 minutes. This includes a 20-minute per-scene fine-tuning phase of Stable-Diffusion, 10 minutes for panorama mesh generation, and 10 minutes for 10000 iterations of 3DGS optimization.

\section{Limitations}
Our method comprises several consecutive steps utilizing multiple pre-trained models, each producing intermediate results that may contain errors. However, there are no mechanisms in place within subsequent steps to correct these errors, leading to error accumulation. For instance, during the depth estimation step, monocular depth estimators encounter challenges with highly reflective and transmissive materials such as glass, and these errors propagate through subsequent steps. 
We rely on pre-trained monocular depth estimators, but increased distance can degrade depth quality. In large indoor settings like long hallways, distant objects will lack detailed geometry.
The mesh is iteratively constructed by back-projecting image-space grid mesh to world space, which means the mesh projected in distance will be sparse with blurry textures when closely inspected.
In addition, this paper assumes that the input panorama captures fundamental room structures to identify camera search space.

\section{Conclusion}
In this paper, we proposed Pano2Room that generates a high-quality 3DGS scene from a single panorama. To achieve this goal, we designed several new modules including a Pano2Mesh module for constructing a panorama's mesh, a panoramic RGBD inpainter designed to generate the occluded content of a scene, an iterative mesh refinement module with camera searching and geometry conflict avoidance strategy to enhance the quality of inpainting, and a Mesh2GS module to boost the quality of novel view synthesis.
Through extensive evaluations on various panorama datasets, we demonstrated that Pano2Room achieves state-of-the-art reconstruction quality in single-panorama novel view synthesis.

% \vspace{-2pt}

\begin{acks}
This work was supported by National Natural Science Foundation of China (Grant No.: 62372015), Center For Chinese Font Design and Research, Key Laboratory of Intelligent Press Media Technology, and State Key Laboratory of General Artificial Intelligence.
\end{acks}

% \pagebreak

% Bibliography
\bibliographystyle{ACM-Reference-Format}
\bibliography{bibliography}

\clearpage

%% figues only 
\pagebreak

\pagebreak

\end{document}